\documentclass[sigconf]{aamas} 

\usepackage{balance} 
\usepackage{graphicx}
\usepackage{amsmath}
\usepackage{dsfont}
\usepackage{subcaption}
\usepackage{wrapfig}
\usepackage{nccmath}
\usepackage{mathtools}
\usepackage{cleveref}
\usepackage{multicol}
\usepackage{multirow}
\usepackage{xcolor}
\usepackage{makecell}
\usepackage{algorithm}  
\usepackage{algorithmicx}  
\usepackage{algpseudocode}
\usepackage[all]{nowidow}

\DeclareMathOperator*{\argmax}{arg\,max}
\DeclarePairedDelimiter{\nint}\lfloor\rceil

\setcopyright{ifaamas}
\acmConference[AAMAS '24]{Proc.\@ of the 23rd International Conference
on Autonomous Agents and Multiagent Systems (AAMAS 2024)}{May 6 -- 10, 2024}
{Auckland, New Zealand}{N.~Alechina, V.~Dignum, M.~Dastani, J.S.~Sichman (eds.)}
\copyrightyear{2024}
\acmYear{2024}
\acmDOI{}
\acmPrice{}
\acmISBN{}

\acmSubmissionID{<<503>>}

\title[]{Context-aware Communication for Multi-agent Reinforcement Learning}

\author{Xinran Li}
\affiliation{
  \institution{The Hong Kong University of Science and Technology}
  \city{Hong Kong}
  \country{China}}
\email{xinran.li@connect.ust.hk}

\author{Jun Zhang}
\affiliation{
  \institution{The Hong Kong University of Science and Technology}
  \city{Hong Kong}
  \country{China}}
\email{eejzhang@ust.hk}

\begin{abstract}
Effective communication protocols in multi-agent reinforcement learning (MARL) are critical to fostering cooperation and enhancing team performance. To leverage communication, many previous works have proposed to compress local information into a single message and broadcast it to all reachable agents. This simplistic messaging mechanism, however, may fail to provide adequate, critical, and relevant information to individual agents, especially in severely bandwidth-limited scenarios. This motivates us to develop context-aware communication schemes for MARL, aiming to deliver personalized messages to different agents. Our communication protocol, named CACOM, consists of two stages. 
In the first stage, agents exchange coarse representations in a broadcast fashion, providing context for the second stage. Following this, agents utilize attention mechanisms in the second stage to selectively generate messages personalized for the receivers. Furthermore, we employ the learned step size quantization (LSQ) technique for message quantization to reduce the communication overhead. 
To evaluate the effectiveness of CACOM, we integrate it with both actor-critic and value-based MARL algorithms. Empirical results on cooperative benchmark tasks demonstrate that CACOM provides evident performance gains over baselines under communication-constrained scenarios. The code is publicly available at https://github.com/LXXXXR/CACOM.
\end{abstract}


\newcommand{\BibTeX}{\rm B\kern-.05em{\sc i\kern-.025em b}\kern-.08em\TeX}

\begin{document}


\pagestyle{fancy}
\fancyhead{}


\maketitle 


\section{Introduction}

Cooperative multi-agent reinforcement learning (MARL) has recently emerged as an exciting research avenue due to its applicability to real-world scenarios. Many of these applications naturally call for multi-agent frameworks, such as resource management~\citep{marl_dis}, package delivery~\citep{marl_delivery}, disaster rescue~\citep{marl_rescue}, and robots control~\citep{marl_robot_swamy2020scaled}. Despite the considerable success of seminal works~\citep{qmix_rashid2020monotonic, maddpg_lowe2017multi, mappo_yu2021surprising} in generalizing reinforcement learning algorithms to multi-agent cases under the centralized training and decentralized execution (CTDE) paradigm~\citep{ctde_kraemer2016multi, ctde_lyu2021contrasting}, further improvement in MARL is hindered by a few prominent obstacles, such as non-stationarity and partial observability~\citep{yuan2023survey,review_coo_oroojlooy2022review}. \looseness=-1

Subsequent research efforts~\citep{CommNet, BiCNet} aim to tackle the non-stationary and partially-observable issues in MARL by incorporating communication into the framework. By integrating local information (observations or histories) with messages from other agents, an agent can gain a more comprehensive understanding of the environment, leading to improved decision-making. Prior methods,  focusing on improving the MARL task performance, mostly adopt continuous messages and broadcasting mechanisms, thus inducing substantial communication overhead.Communication channels typically face severe bandwidth constraints~\citep{mas_bandwidth_huang2016distributed}, making it difficult to directly deploy previous approaches to real-world multi-agent systems (MASs). Therefore, it is crucial to develop MARL communication protocols that can utilize the limited bandwidth more efficiently. \looseness=-1

\begin{figure*}[t]
  \centering
  \includegraphics[scale=0.23]{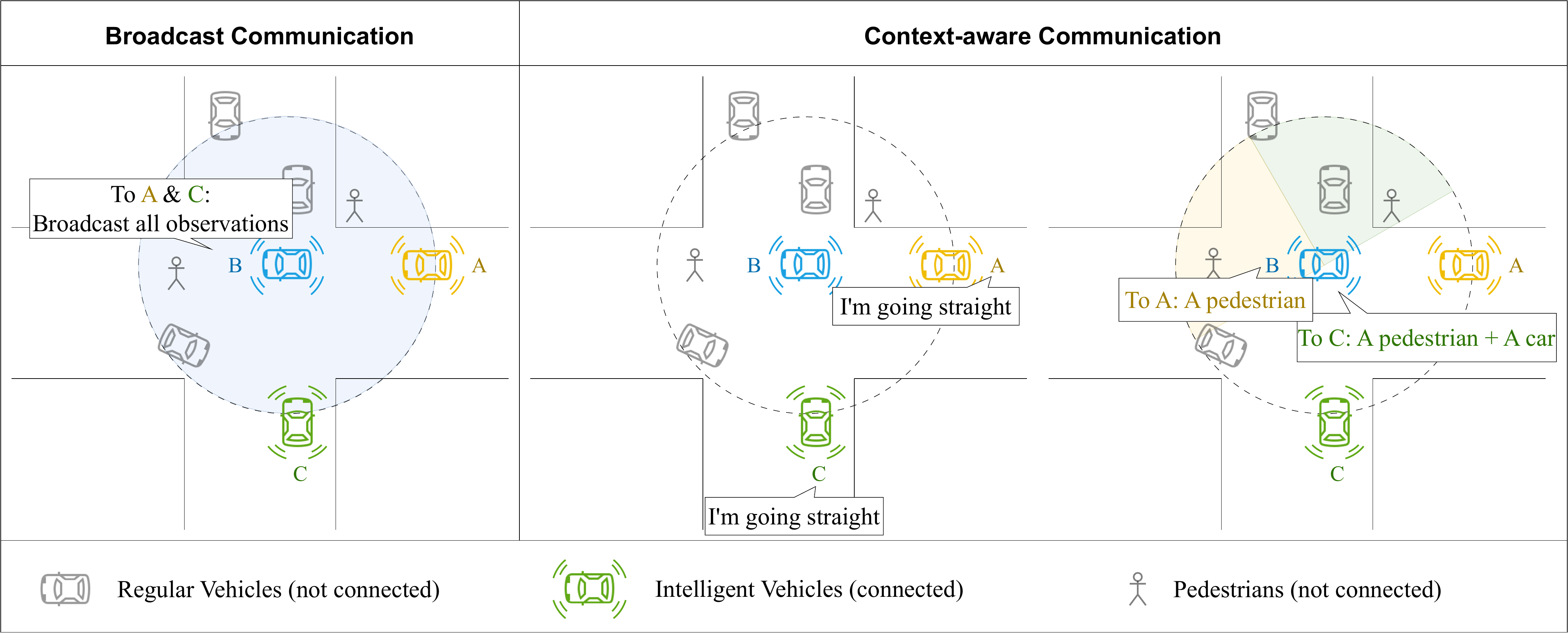} 
  \caption{An illustrative example in a busy traffic junction. Under the broadcasting communication scheme, without knowledge of agent A and C's intentions, agent B needs to broadcast all its observations, which will result in heavy communication overhead. In contrast, when adopting context-aware communication, agent A and C first convey their local information in short context messages. Then agent B generates personalized messages for A and C based on the context messages from the previous stage. In this way, more context-relevant messages are provided for decision making with much lower communication overhead. }
  \label{fig:example}
  \Description{An illustrative example in a busy traffic junction. Under the broadcasting communication scheme, without knowledge of agent A and C's intentions, agent B needs to broadcast all its observations, which will result in heavy communication overhead. In contrast, when adopting context-aware communication, agent A and C first convey their local information in short context messages. Then agent B generates personalized messages for A and C based on the context messages from the previous stage. In this way, more context-relevant messages are provided for decision making with much lower communication overhead. }
\end{figure*}

In this work, we consider an MAS with a limited communication budget. Ideally, a communication protocol that operates well under a tight budget shall only allow transmitting information that is useful for the target receiver, i.e., the message generation shall be context-aware. This makes the broadcast communication protocol a suboptimal choice, as different agents likely need different information. But the question remains, i.e., how to identify the pieces of information worth transmitting between two particular agents? 
Note that MARL aims to solve a decision-making problem at its core, therefore it is up to the receivers to determine what information is relevant and helpful for decision making. This motivates us to develop a receiver-centric communication protocol, in which the receiver initiates the communication by providing the sender with the ``context'' before the sender can send context-aware messages. We compare ideas of the traditional broadcasting scheme and context-aware communication (CACOM) using an illustrative example in Figure~\ref{fig:example}. In the first stage of our proposed CACOM, each agent broadcasts a very short context message conveying coarse local information. In the second stage, agents selectively send longer, personalized messages to specific peers based on the context messages they received in the first stage. It is noteworthy that similar communication strategies among humans have been successfully utilized in the fields of marketing~\citep{idea_market_ansari2003customization} and healthcare~\citep{idea_health_kreuter2013tailoring}, where they are termed as customized communication. 

To aid personalized message generation, we treat the entities in agents' observation as tokens and employ an attention-based architecture to encode features and generate personalized messages. This approach enables the network to learn which parts of the features are most important for the intended receiver. Furthermore, we quantize the messages through the LSQ technique ~\citep{lsq_esser2020learned} to make the communication blocks compatible to digital communication systems.  
Since CACOM is agnostic to specific MARL algorithms, we combine it with MADDPG~\citep{maddpg_lowe2017multi} and QMIX~\citep{qmix_rashid2020monotonic} to evaluate its efficacy on cooperative multi-agent benchmarks in communication-constrained scenarios. Specifically, we conduct experiments on two scenarios in multi-agent particle environment (MPE)~\citep{maddpg_lowe2017multi} and four maps in the StarCraft multi-agent challenge (SMAC)~\citep{smac_samvelyan19smac}. The results showcase the superior performance of CACOM in comparison with baselines. 
Our contributions are summarized as follows: 
\begin{itemize}
    \item
    We investigate a realistic multi-agent system with limited communication resources and propose a context-aware communication protocol in MARL, namely CACOM, to efficiently utilize the limited communication budgets.
    \item We leverage various attention blocks to generate personalized messages based on the senders' and receivers' local information, and we incorporate learned step size quantization (LSQ) to ensure digital communication while keeping the overall network differentiable. 
    \item Through adequate experimentation on cooperative multi-agent benchmarks with limited communication budgets, CACOM has been demonstrated to be effective and outperform traditional broadcast communication protocols.
\end{itemize}


\section{Related Work}

In recent years, there has been a surge of interest in the field of MARL, particularly in the context of CTDE~\citep{ctde_kraemer2016multi}. Numerous algorithms have been developed to support cooperative MARL, including QMIX~\citep{qmix_rashid2020monotonic}, MAPPO~\citep{mappo_yu2021surprising}, and MADDPG~\citep{maddpg_lowe2017multi}.
To mitigate the impact of partially observablility and promote cooperation, communication has been incorporated into MASs~\citep{CommNet, BiCNet}. 
Notably, by applying attention mechanisms and graph neural networks (GNNs), TarMAC~\citep{Tarmac}, DICG~\citep{dicg} and DGN~\citep{DGN} learn local embeddings and broadcast the messages to all the reachable agents. Subsequent methods propose to further improve the performance from two aspects: the sender side and the receiver side. From the sender side, MAIC~\citep{maic_yuan2022multi} and ToM2C~\citep{tom2c_wang2021tom2c} utilize teammate modeling to generate incentive messages based on identities of the receivers. The other line of research aims to develop more delicate aggregation schemes at the receiver side to utilize the received messages more efficiently, leading to algorithms such as G2A~\citep{g2a} and MASIS~\citep{masis_guan2022efficient}. However, most of these works focus more on enhancing the MARL task performance, with less emphasis on controlling the communication costs, which could result in prohibitive communication overhead in real-world systems. 

There have been methods, e.g., ATOC~\citep{atoc}, IC3~\citep{ic3} and I2C~\citep{i2c}, trying to reduce the communication overhead by applying local gating mechanisms to dynamically prune the communication links among agents. However, since the messages are continuous, it remains unknown whether these methods can work well in systems with extremely limited communication budgets. In contrast, DIAL~\citep{DIAL} and vector quantization-based methods~\citep{vq_MARL_liu2022adaptive} generate discrete messages directly, while NDQ~\citep{ndq_wang2019learning} and TMC~\citep{tmc_zhang2020succinct} also designs messages in a compact way. Nevertheless, the expressiveness of discrete messages will be limited by the communication budget with broadcasting communication schemes. Furthermore, ETC~\citep{etc_hu2021event}, VBC~\citep{VBC} and MBC~\citep{MBC_sparse_comm} have proposed event-triggered communication to reduce the communication frequency and address the communication constraints. This line of research focuses on optimizing when transmissions shall occur, thus is orthogonal to our work and can potentially be combined with our work to achieve better performance with lower communication budgets.

In this work, we explicitly account for the effect of limited communication budgets in MASs. We utilize the LSQ technique to ensure that messages are discrete and therefore applicable to communication-constrained scenarios. Moreover, we leverage the context-aware communication protocol to generate optional links and personalized messages, enabling more efficient use of communication resources. Compared with previously proposed incentive messages~\citep{tom2c_wang2021tom2c, maic_yuan2022multi} based on identities instead of specific context, CACOM can better address environment dynamics and generate more compact messages for receivers. Comprehensive comparisons with existing methods will illustrate the effects of limited communication resources, as well as the importance of sending personalized messages to agents. 


\section{Background}
\textbf{Decentralized Partially Observable Markov Decision Process (Dec-POMDP): }
In this work, we consider a fully cooperative partially observable multi-agent task, which can be modeled as a decentralized partially observable Markov decision process (Dec-POMDP)~\citep{pomdp_oliehoek2016concise}. The Dec-POMDP is defined by a tuple $\mathcal{M} = \langle \mathcal{S}, A, P, R, \Omega, O, n, \gamma \rangle$, where $n$ denotes the number of agents and $\gamma \in (0, 1]$ is the discount factor that balances the trade-off between immediate and long-term rewards. 

At each timestep $t$, when the environment state is $s \in \mathcal{S}$, each agent $i$ receives a local observation $ o_i \in \Omega$ drawn from the observation function $O(s, i)$. Before taking any actions, agents communicate with each others and exchange local information. Subsequently,  each agent $i$ follows its local policy $\pi_i$ to select an action $a_i \in A$ based on its local information and information obtained from communication. Individual action decisions will then form a joint action $\boldsymbol{a} \in A^n$, which results in a state transition to the next state $s' \sim P(s'| s, \boldsymbol{a})$ and a global reward $r = R(s, \boldsymbol{a})$. Each agent keeps a local action-observation history denoted as $h_i  \in (\Omega \times A)$. The team objective is to learn the policies such that the expectation of discounted accumulated reward $G_t = \sum_t \gamma^t r^t$ is maximized. 

\textbf{Centralized Training and Decentralized Execution (CTDE): } 
The CTDE~\citep{ctde_kraemer2016multi, ctde_lyu2021contrasting} paradigm in MARL, which combines the advantages of centralized training for improved coordination and decentralized execution for enhanced scalability, has emerged as a popular approach for solving complex problems in MASs.
Recently, much research effort has been made to generalize single-agent RL algorithms to multi-agent cases under the CTDE paradigm, resulting in a variety of concrete MARL algorithms. In this work, we build our proposed communication blocks on top of both actor-critic algorithm MADDPG~\citep{maddpg_lowe2017multi} and value-based algorithm QMIX~\citep{qmix_rashid2020monotonic} for different environments, aiming to showcase the feasibility of further improving the learning performance of these algorithms when communication is allowed. Training objectives and details of these MARL algorithms are provided in Supplementary~\ref{sec: supp_bg}.

\textbf{Learned Step Size Quantization (LSQ):} When developing communication blocks for MARL algorithms, the desideratum is to have discrete messages, which can be transmitted in real-world digital communication systems. At the same time, we also want to ensure the end-to-end differentiable property of the overall network architecture. To meet such requirements without impeding the performance, we migrate quantization techniques in low-precision networks into MARL network architecture design.
In particular, we choose LSQ~\citep{lsq_esser2020learned} due to its simplicity of implementation and its ability to maintain high accuracy in the low-bit precision networks across a wide range of network architectures on the benchmark dataset ImageNet~\citep{imagenet_deng2009imagenet}. In the LSQ algorithm, the quantizer step size is treated as a learnable parameter rather than a fixed value, which allows more flexibility.

\section{Context-aware Communication for MARL}
In this section, we first introduce CACOM, the two-stage communication protocol we designed to enable context-aware communication in MARL. We then detail the network architecture to accommodate CACOM. Lastly, we provide an overview of the training objective and the techniques we employed for training. For the sake of readability, we only discuss the implementation on top of the QMIX algorithm in this section, but it can be easily generalized to the MADDPG algorithm. Pseudo codes for CACOM can be found in Supplementary~\ref{sec: supp_impl}.

\subsection{Communication Protocol Design}

The objective of MARL is for agents to choose actions locally so that the joint action will lead to a better team reward. From an agent's perspective, the communication messages it receives can augment its local information, thereby assisting action decisions. To ease the explanation, we call this agent receiving information the \textit{helpee} and the agent providing information the \textit{helper}. Note that in our proposed method, agents are homogeneous, therefore each agent behaves both as a helpee and a helper at the same time. 

\begin{figure}
  \includegraphics[scale=0.9]{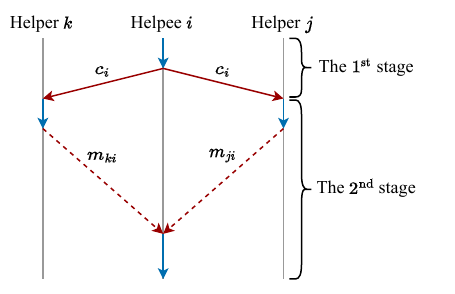}
  \centering
  \caption{Illustration of the CACOM protocol from the helpee agent $i$'s perspective. The blue arrows denote local processing and the red arrows denote communication. At each timestep, agent $i$ first broadcasts a context message $c_i$ to all its peers agent $k$ and $j$. Then, after local processing, agents $k$ and $j$ decide whether to reply and what to send, respectively. }
  \label{fig:protocol}
  \Description{Illustration of the CACOM protocol from the helpee agent $i$'s perspective. The blue arrows denote local processing and the red arrows denote communication. At each timestep, agent $i$ first broadcasts a context message $c_i$ to all its peers agent $k$ and $j$. Then, after local processing, agents $k$ and $j$ decide whether to reply and what to send, respectively. }
\end{figure}

\begin{figure*}[t]
  \centering
  \includegraphics[scale=0.9]{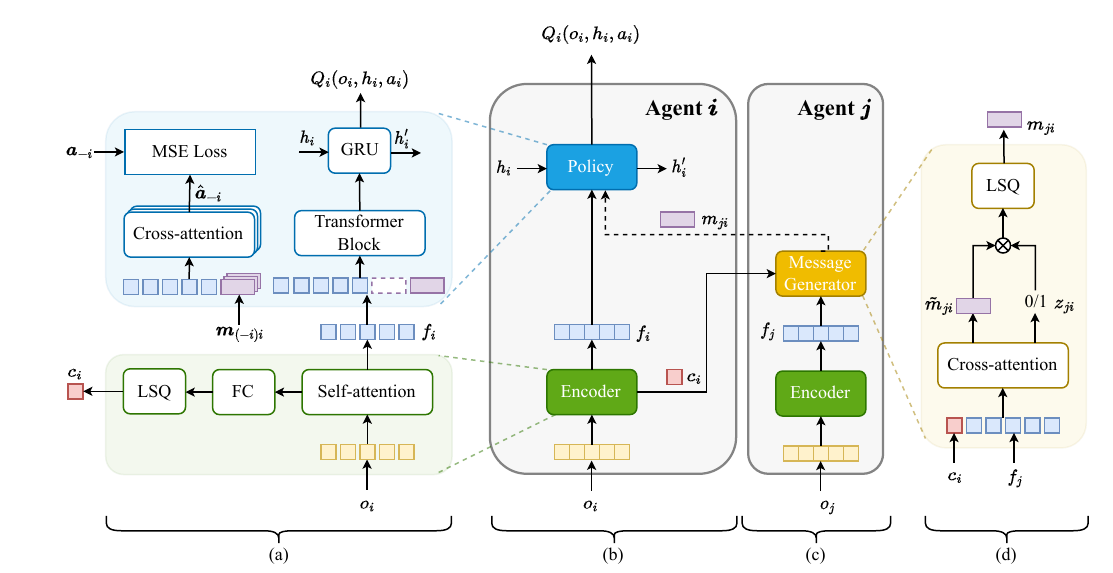} 
  \caption{Network architecture for CACOM. (a) Helpee's feature encoder and policy network. (b) Overall architecture for a helpee agent $i$. (c) Overall architecture for a helper agent $j$. (d) Helper's message generator. }
  \label{fig:arch}
  \Description{Network architecture for CACOM.}
\end{figure*}

Thinking from a helpee's perspective, to improve the overall performance, different helpees may need different information to make more sensible decisions. If the helpers send messages in a broadcast fashion, to accommodate the various need for different helpees, this single message has to contain ample information, inevitably inducing heavy communication overhead. At the same time, the helpees will end up being flooded with a massive amount of irrelevant information~\citep{g2a}, which can even impede the decision making process. To this end, we propose a two-stage communication protocol, namely CACOM, in which the messages are personalized to the helpees' need.

The proposed CACOM is demonstrated in Figure~\ref{fig:protocol}. In the first stage, the helpee,  agent $i$, broadcasts a short \textit{context message} $c_i$ to all its peers in the system. $c_i$ is a coarse representation of the local information, ideally indicating the available information at agent $i$. Upon receiving the messages from the first stage, a potential helper agent $j$ first decides locally whether it can help the helpee. If agent $j$ decides not to be the helper of agent $i$, then no message will be sent from agent $j$ to agent $i$ in the second stage. Otherwise, agent $j$ will act as a helper and generate a \textit{personalized message} $m_{ji}$ based on the context message $c_i$ from the helpee as well as agent $j$'s local information. After receiving all the messages from the second stage, the helpee $i$ aggregates the information and chooses an action to execute. \looseness=-1

\subsection{Agent Network Design}\label{sec:network}

To implement CACOM, we designed the agent network architecture as depicted in Figure~\ref{fig:arch}. It is composed of three blocks: an encoder, a message generator and a policy network. In particular, the encoder at the helpee agent $i$ takes local observation $o_i$ as input, and outputs the local feature $f_i$ and the context message $c_i$ for the first communication stage. Upon receiving $c_i$, the message generator at the helper agent $j$ determines locally whether to assist agent $i$ and generates the personalized message $m_{ji}$ for the second communication stage if needed. After the second communication stage, the policy network at the helpee agent $i$ aggregates the received personalized messages $\boldsymbol{m}_{(-i)i}$ and the local feature $f_i$ to calculate the local value.

\textbf{Attention Architecture to Generate Personalized Messages:}
As discussed before, we aspire for the message generator at the helper side to produce a personalized message for the helpee based on the received context information. This is achieved by the attention-based blocks which encode entities from the observation respectively and generate personalized messages. 

At the helpee agent $i$'s side, the encoder treats the observation as tokens of entities, where entities can be ego states, enemy states, teammate states or environment states. By applying self-attention to tokens of entities, the local encoder generates a local feature $f_i \in \mathbb{R}^{m \times d_f} $ ($m$ denotes the number of entities and $d_f$ denotes the encoding dimension for each entity) and a context message $c_i$ for broadcasting. At the helper agent $j$'s side, the message generator utilizes the cross-attention mechanism and takes the received message $c_i$ and the local feature as input and generates the personalized message $m_{ji}$: 
\begin{align}
    \tilde{m}_{ji} &= \textrm{softmax} \left(\frac{q_i^T k_j }{\sqrt{d_k}} \right) v_j, \\
    m_{ji} &= \textrm{LSQ}\left( \tilde{m}_{ji} \cdot z_{ji} \right) \label{eq:reply}, 
\end{align}

where $k_j = W_k f_j \in \mathbb{R}^{d_k \times m}, q_i = W_q c_i \in \mathbb{R}^{d_k \times 1}, v_j = W_v f_j  \in \mathbb{R}^{d_m \times m}$, $d_m$ denotes the message dimension for the second stage communication and the $\textrm{softmax}(\cdot)$ is calculated along the $m$ dimension. $z_{ji}$ is a one-bit value denoting whether to generate a personalized message to $i$, and we delay further details for Equation~\ref{eq:reply} later in this section.

After receiving context-aware messages from all the helpers, the helpee agent $i$ uses a transformer block to aggregate the received messages $\boldsymbol{m}_{(-i)i}$ and its own local feature $f_i$. This is followed by a gated recurrent unit (GRU) block, which considers historical information $h_i$ and outputs a local value: 
\begin{align}
    Q_i(o_i, h_i, \cdot; \theta) = \mathcal{F} \left( f_i, h_i, \boldsymbol{m}_{(-i)i}; \theta \right),
    \label{eq: aggregation}
\end{align}
where $\mathcal{F}$ denotes the policy function.

\textbf{Quantization to Reduce Communication Overhead:} 
As practical systems adopt digital communication, it is essential to use a differentiable quantizer to digitize the messages. 
To this end, we adopt LSQ as the quantizer for both communication stages, which offers a simple implementation and high accuracy for gradient estimation. Following the LSQ algorithm, we have
\begin{align}
    m_{ji} = \nint{\textrm{clip}(\tilde{m}_{ji} \cdot z_{ji}, - 2^{b-1}, 2^{b-1}-1)} \cdot s_\theta,
\end{align}
where $\textrm{clip}(\cdot, r_1, r_2)$ returns the input with the values below $r_1$ set to $r_1$ and the values above $r_2$ set to $r_2$, $\nint{\cdot}$ rounds input to the nearest integer, $b$ denotes number of bits and $s_\theta$ denotes the learnable step size.
This approach allows for digital communication while preserving end-to-end differential properties.

\textbf{Gating Mechanism to Prune Unnecessary Links:} 
To further reduce the communication overhead and eliminate unncessary communication links, we utilize the gating mechanism~\citep{atoc} to prune the links for the second communication stage dynamically. The context messages from the first  communication stage can naturally serve as indications on how to prune the links in the second communication stage. Specifically, we train a local binary classifier at the potential helper agent $j$'s side to determine whether to send a message at the second stage to a particular helpee $i$, based on the helper's local feature $f_j$ and the context message from the helpee $c_i$. 
\begin{align}
    z_{ji} &= \mathds{1} \left[\mathcal{G}(f_j, c_i; \theta_\mathcal{G}) > 0.5 \right], \\
    \mathcal{G}(f_j, c_i; \theta_\mathcal{G}) &= \textrm{sigmoid}\left(\textrm{FC}\left(\frac{q_i'^T k_j' }{\sqrt{d_k}} \right)\right),
    \label{eq:local_gate}
\end{align}

where $k_j' = W_k' f_j \in \mathbb{R}^{d_k \times m}, q_i' = W_q' c_i \in \mathbb{R}^{d_k }$, $\mathds{1} [\cdot]$ is the indicator function, $\textrm{FC}(\cdot)$ is the fully-connected (FC) layer and $\theta_\mathcal{G}$ is the parameters for the gate.

We train the binary classifier in a self-supervised manner, leveraging the centralized value function to generate pseudo labels $y_{ji}$ for local gates during training.
In particular, we compare the global value when $z_{ji} = 1$ with the one when $z_{ji} = 0$:
\begin{align}
    y_{ji} = \mathds{1} \left[ 
    Q_{\text{tot}}(\boldsymbol{o}, \boldsymbol{h}, \boldsymbol{a}_{-i}, a_i^1; \theta) - Q_{\text{tot}}(\boldsymbol{o}, \boldsymbol{h}, \boldsymbol{a}_{-i}, a_i^0; \theta) > T
    \right],
    \label{eq:psuedo_labels}
\end{align}

where 
\begin{align}
    a_i^0 &= \argmax_a Q_i^0(o_i, h_i, a), \\
    a_i^1 &= \argmax_a Q_i^1(o_i, h_i, a), \\
    Q_i^0(o_i, h_i, \cdot ; \theta) &= \mathcal{F} \left(f_i, h_i, \boldsymbol{m}_{(-i)i}; \theta \right)|_{z_{ji} = 0}, \\
    Q_i^1(o_i, h_i, \cdot; \theta) &= \mathcal{F} \left( f_i, h_i, \boldsymbol{m}_{(-i)i}; \theta \right)|_{z_{ji} = 1},
\end{align}
 where $T$ is a threshold. 
 
 In this way, we prune the communication links that fail to contribute to the global value.

\subsection{Training Details}

Following the QMIX algorithm, parameters $\theta$ are updated by the TD loss:
\begin{align}
    \mathcal{L}_{\textrm{TD}}(\theta) &= \mathbb{E}_{(\boldsymbol{o}, \boldsymbol{h}, \boldsymbol{a}, r, \boldsymbol{o}', \boldsymbol{h}') \sim \mathcal{D}} \left[ \left(y - Q_{\text{tot}}(\boldsymbol{o}, \boldsymbol{h}, \boldsymbol{a}; \theta) \right)^2 \right], \\
    y &= r + \gamma \max_{\boldsymbol{a}'} Q_{\text{tot}}(\boldsymbol{o}', \boldsymbol{h}', \boldsymbol{a}'; \theta^-), 
\end{align}
where $y$ is the target and $\theta^-$ is the parameters for the target network. 

Moreover, since the two-stage communication blocks may enlarge the policy space, making the optimization even more challenging, we introduce an auxiliary loss to prevent the communication from collapsing (i.e., not transmitting any helpful information). We design the auxiliary task to predict the helpers' local values at the helpee's side utilizing the local feature and received personalized messages. Specifically, in addition to the TD loss, we minimize an auxiliary loss given as the mean squared error (MSE) of predictions:
\begin{align}
    \mathcal{L}(\theta_\mathcal{P}) = \mathbb{E}_{(h_j, f_i, m_{ji}) \sim \mathcal{D}} \left[ (Q_j(h_j, a; \theta) - \mathcal{P}(f_i, m_{ji}; \theta_\mathcal{P}))^2 \right],
\end{align}
where $\theta_\mathcal{P}$ denotes the parameters for the communication blocks and the predictor, and $\mathcal{P}$ denotes the predictor function. Note that this auxiliary task is only used to regularize the parameters training in the communication blocks, and the predictions will not be utilized to assist helpee's action decisions.

As previously mentioned, we train the local gate in a self-supervised manner, with synthetic labels generated during training. The local gate $\mathcal{G}$ is trained alternately with the rest of the network. To prevent all communication links from being cut off at the beginning, we disable gate training during the early stages. The loss function for the local gates is defined as follows:
\begin{align}
    \mathcal{L}(\theta_\mathcal{G}) = \mathbb{E}_{(c_i, f_j) \sim \mathcal{D}} &\left[y_{ji} \cdot \log(\mathcal{G}(f_j, c_i ; \theta_\mathcal{G}))\right. \nonumber\\
    &- \left.(1 - y_{ji}) \cdot \log(1 - \mathcal{G}(f_j, c_i; \theta_\mathcal{G}))\right],
\end{align}
where $y_{ji}$ is generated following Equation ~\ref{eq:psuedo_labels}.

To improve the sample efficiency as well as lower the model complexity, we adopt parameter sharing among agents.

\section{Experiments}\label{sec: exp}

\begin{figure*}[t]
\centering
\begin{subfigure}[t]{.23\linewidth}
    \centering
    \includegraphics[width=\linewidth]{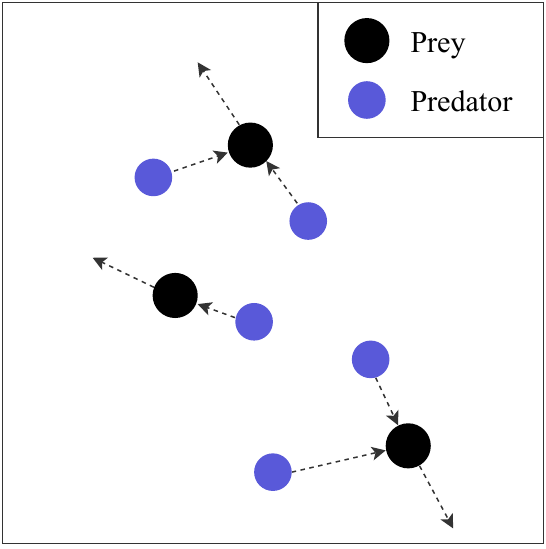}
    \caption{Predator-prey}
\end{subfigure}
\begin{subfigure}[t]{.23\linewidth}
    \centering
    \includegraphics[width=\linewidth]{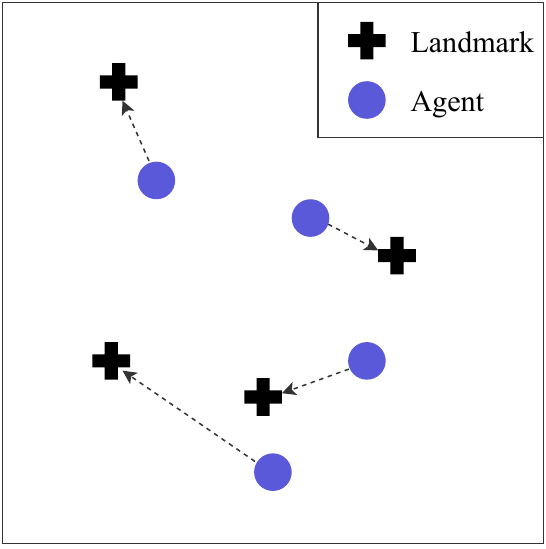}
    \caption{Cooperative navigation}
\end{subfigure}
\begin{subfigure}[t]{.37\linewidth}
    \centering
    \includegraphics[width=\linewidth]{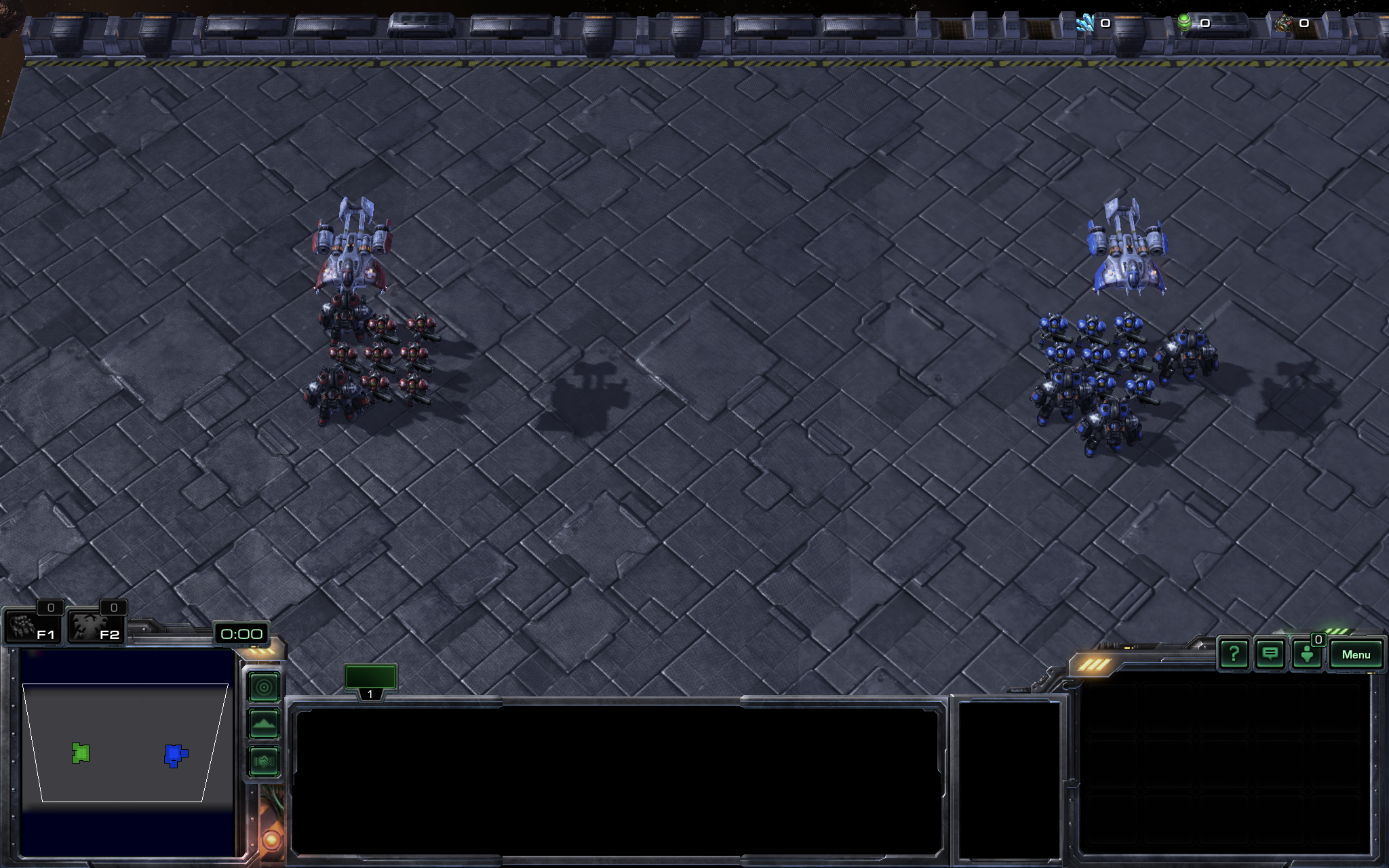}
    \caption{SMAC}
\end{subfigure}
\caption{Multi-agent environments.}
\label{fig:environments}
\Description{Multi-agent environments including SMAC and MPE.}
\end{figure*}

In this section, we evaluate our proposed CACOM protocol on three multi-agent cooperative benchmark tasks shown in Figure~\ref{fig:environments}: predator-prey (PP) and cooperation navigation (CN) in MPE \citep{maddpg_lowe2017multi}, and SMAC \citep{smac_samvelyan19smac}. As CACOM is agnostic to specific MARL algorithms, we build the proposed communication blocks on top of both value-based algorithms and actor-critic algorithms to test out the applicability. In particular, we follow the literature and use the MADDPG algorithm for the continuous tasks of MPE, and the QMIX algorithm for the discrete task SMAC.

Furthermore, real-world communication systems cannot send continuous signals, so we allocate different communication budgets (in the number of bits per link per timestep) for each scenario based on the observation and action spaces. For CACOM, we use the communication overhead from both stages combined together as the overall communication overhead. We then test out the average rewards or success rates to compare different algorithms. 
For evaluation, all experiments are averaged with five random seeds and the shaded areas refer to the 95\% confidence interval. Details of the network architectures and training hyperparameters are given in Supplementary~\ref{sec: supp_impl}.

\subsection{MPE}


PP is a cooperative MARL benchmark task where the goal for agents (predators) is to capture the moving preys. We set the number of predators to be 10 and the number of preys to be 4 in this experiment. 
CN requires the agents to occupy stationary landmarks. 
In this experiment, we set both the number of agents and the number of landmarks to be 8. The world for both scenarios is a continuous space (as opposed to a grid world). The communication budget is set to $24$ bits per link per timestep in PP and $32$ bits per link per timestep in CN for all algorithms. We include more details on the environmental settings in Supplementary~\ref{sec: supp_env}. 

\textbf{MARL Algorithms and Baselines:} For this environment, we build CACOM on top of the MADDPG algorithm and then evaluate it against MADDPG, TarMAC + LSQ, I2C + LSQ, and MAIC + LSQ. 
The baselines, except MADDPG, do not generate discrete messages, and directly pruning the messages would result in unsatisfactory outcomes. To address this issue, we augment TarMAC, I2C, and MAIC with LSQ blocks for fair comparison. In addition, we introduce extra FC layers before and after the communication for MAIC and I2C to enable them to operate within the communication budget. The hyperparameters in both baselines and CACOM were tuned to obtain the best performance.

\begin{table}[h]
\centering
\caption{Gate Pruning Results for MPE}
\begin{tabular}{ccc}
\toprule
scenarios & \makecell{second-stage links \\ pruned ratio} & \makecell{communication budget \\ occupied ratio} \\ \midrule
PP  &  $36.33 (\pm 12.73)\%$ &  $75.78 (\pm 8.48)\%$ \\ 
CN &  $25.88 (\pm 15.81)\%$ &  $80.59 (\pm 11.85)\%$ \\ 
\bottomrule
\end{tabular}
\label{table: gate_MPE}
\end{table}

\begin{figure*}[t]
\centering
\begin{subfigure}[c]{.37\linewidth}
    \centering
    \includegraphics[width=\linewidth]{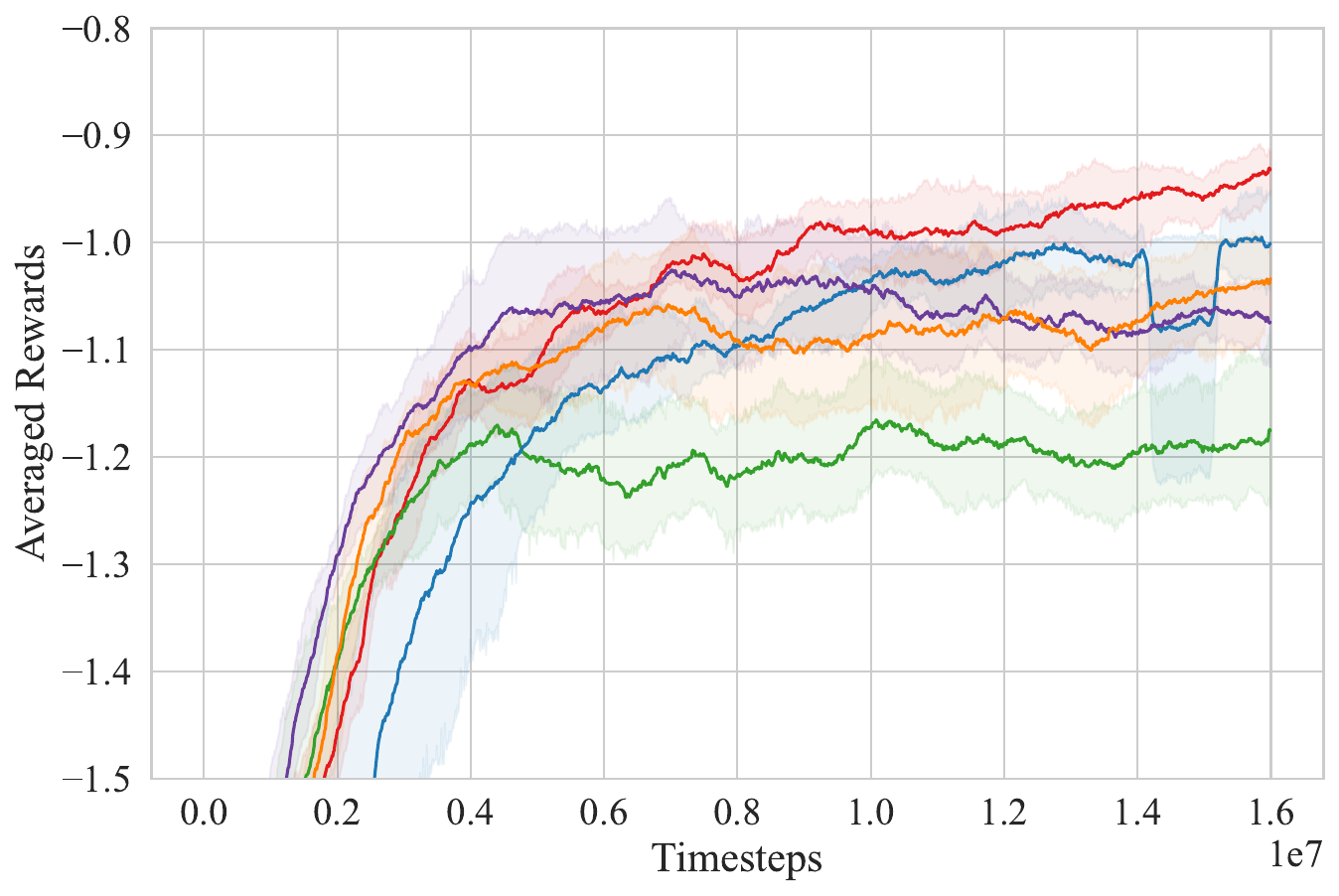}
    \caption{Predator-prey}
    \label{fig:mpe_pp}
\end{subfigure}
\begin{subfigure}[c]{.37\linewidth}
    \centering
    \includegraphics[width=\linewidth]{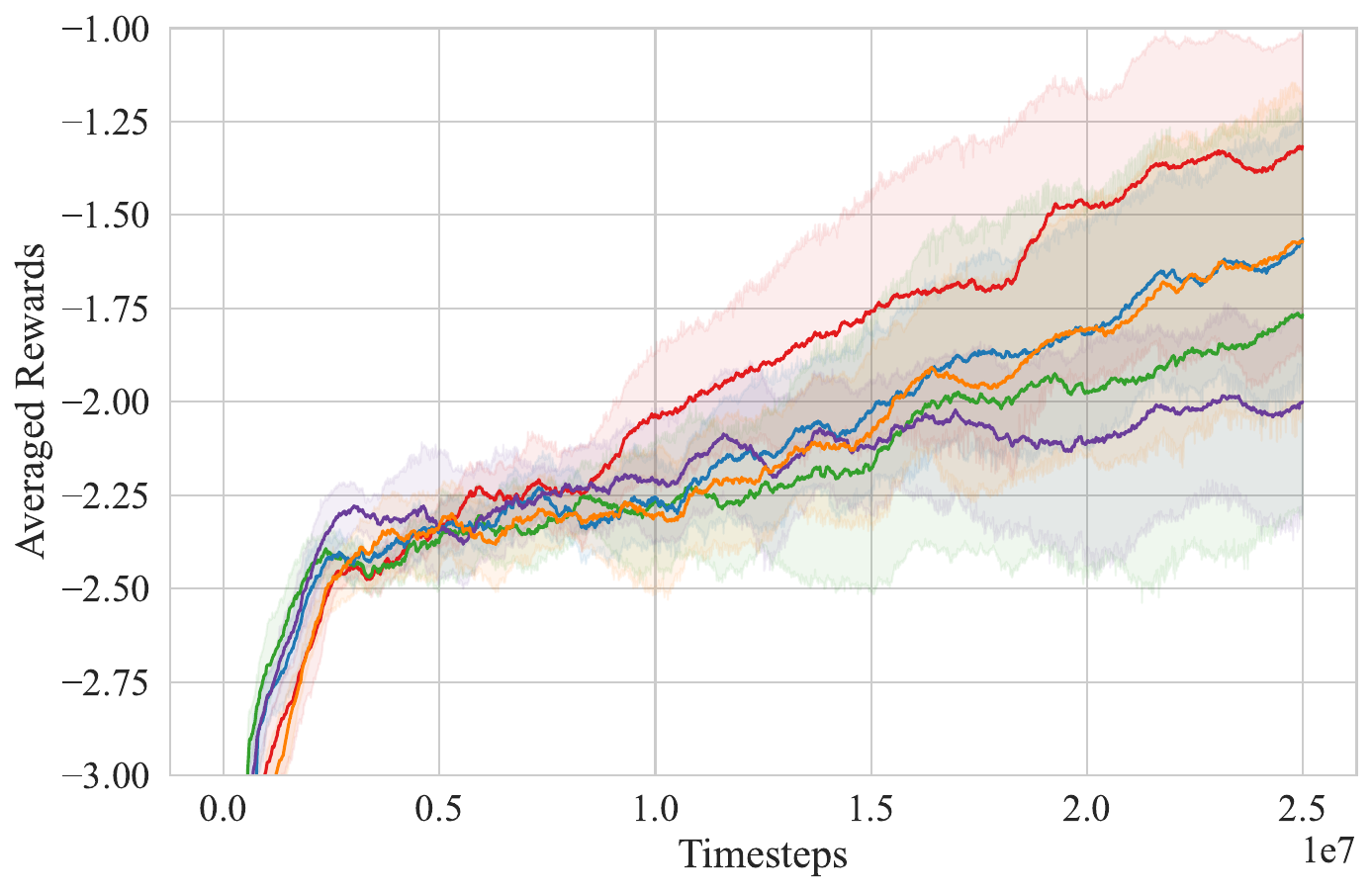}
    \caption{Cooperative navigation}
    \label{fig:mpe_cn}
\end{subfigure}
\begin{subfigure}[c]{.15\linewidth}
    \centering
\includegraphics[width=\linewidth]{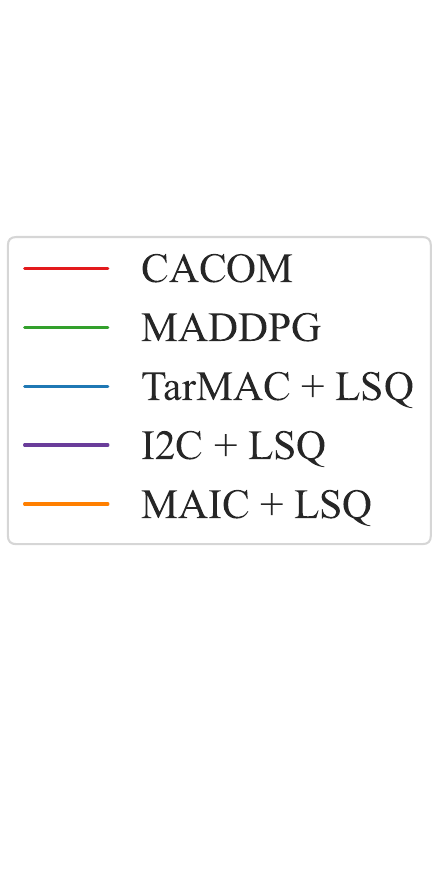}

\end{subfigure}

\caption{Performance comparison with baselines on MPE benchmarks.
}
\label{fig:results_mpe}
\Description{}{Performance comparison with baselines on MPE benchmarks, where CACOM achieves the best performance.
}
\end{figure*}

\textbf{Results:} We show the performance of CACOM and baselines on MPE benchmarks in Figure~\ref{fig:results_mpe} and the corresponding gate pruning results in Table~\ref{table: gate_MPE}. It can be observed from Figure~\ref{fig:results_mpe} that CACOM outperforms all the baselines in both tasks, which shows the superiority of context-aware communication under low communication budgets. \looseness=-1

In both tasks, most communication-enhanced methods exceed the performance of MADDPG, even under severe communication constraints. Compared to the performance in the corresponding original papers~\citep{Tarmac, i2c, maic_yuan2022multi}, where continuous messages are transmitted, we only observe modest performance degradation in these baseline communication methods. This shows great potential in applying quantization methods in MARL to reduce communication overhead. However, in some cases (e.g., I2C in Figure~\ref{fig:mpe_cn}), we see that inappropriate communication protocol design may harm the overall performance by providing messages that fail to assist action decisions at the helpee's side. The comparison between CACOM and other communication methods indicates the superior performance of context-aware communication. And this could attribute to the high flexibility of message generation in such a non-broadcast communication protocol and personalized messages based on the helpee's context. \looseness=-1

\begin{figure*}[t]
\centering
\begin{subfigure}[t]{.49\linewidth}
    \centering
    \includegraphics[width=\linewidth]{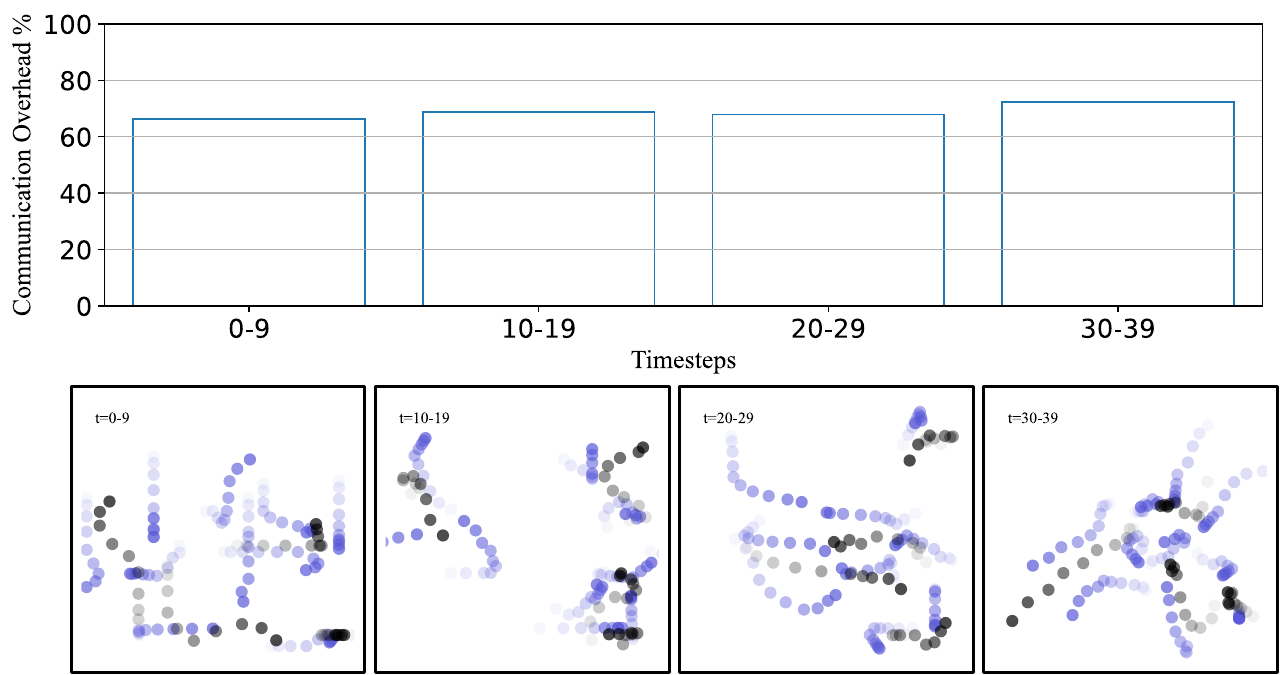}
    \caption{Predator-prey}
    \label{fig:pp_vis}
\end{subfigure}
\begin{subfigure}[t]{.49\linewidth}
    \centering
    \includegraphics[width=\linewidth]{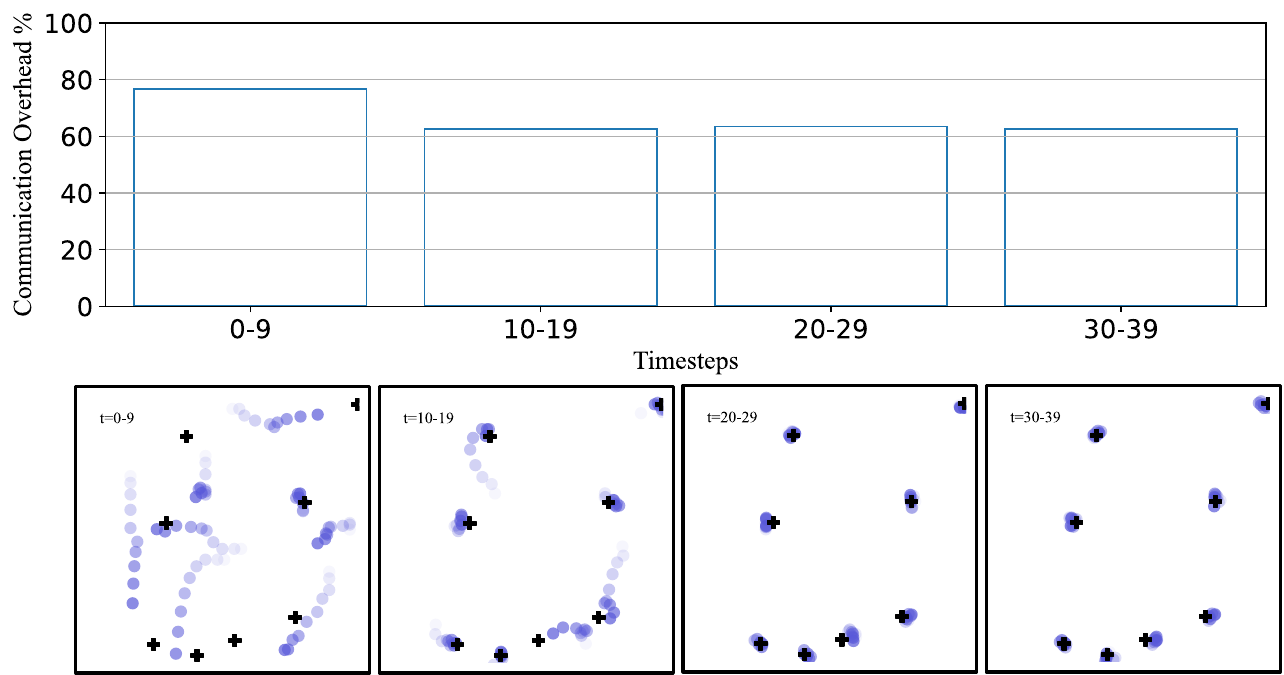}
    \caption{Cooperative navigation}
    \label{fig:cn_vis}
\end{subfigure}

\caption{Visualization of policies in MPE. On the top, we show the communication overhead versus timesteps, in which the communication overhead measures the percentage of the utilized communication budgets. On the bottom, we stack ten timesteps in one figure. The markers correspond to Figure~\ref{fig:environments}, and darker dots correspond to more recent timesteps.}
\label{fig:results_mpe_vis}
\Description{Visualization of policies in MPE. On the top, we show the communication overhead versus timesteps, in which the communication overhead measures the percentage of the utilized communication budgets. On the bottom, we stack ten timesteps in one figure. The markers correspond to Figure~\ref{fig:environments}, and darker dots correspond to more recent timesteps.}
\end{figure*}

\textbf{Visualization:} In Figure~\ref{fig:results_mpe_vis}, we visualize the policies alongside the corresponding communication overhead in two instances of PP and CN, respectively. We see that for instance shown in the Figure~\ref{fig:pp_vis}, agents divide into subgroups and surround the preys. The communication overhead remains roughly the same because the observations are constantly changing as the episode progress. 
In Figure~\ref{fig:cn_vis}, at the early stage of the episode, each agent's target landmark is unclear, therefore more communication links are established among agents. At a later stage, when each agent already occupies a landmark, the communication overhead drops due to the decreasing need. Overall, it shows that the policies learnt by CACOM demonstrate strong cooperation among agents, and the local gate can dynamically prune the unnecessary links according to the context.  \looseness=-1


\subsection{SMAC}

We further evaluate CACOM on SMAC~\citep{smac_samvelyan19smac, maic_yuan2022multi} across four super hard maps: MMM3, 1c3s5z\_vs1c3s6z, 27m\_vs\_30m and corridor. The communication budget is set to be $2 \times \textrm{n\_actions}$ per link per timestep for each scenario. 

\textbf{MARL Algorithms and Baselines: }
For this environment, we build CACOM on top of the QMIX algorithm and evaluate it against QMIX, NDQ, TMC and MAIC + LSQ. Similarly to the experiments in MPE, we augment MAIC with LSQ blocks for quantization. Since NDQ and TMC already consider the communication bandwidth and generate messages in compact ways, we directly adopt the original implementation. We use the hyperparameters in PyMARL~\citep{smac_samvelyan19smac} for both baselines and CACOM.

\begin{figure*}[t]
\centering
\begin{subfigure}[t]{\linewidth}
    \centering
    \includegraphics[width=0.7\linewidth]{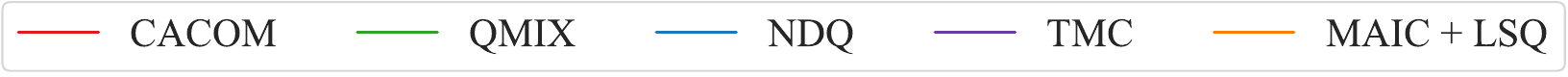}
\end{subfigure}
\begin{subfigure}[t]{.32\linewidth}
    \centering
    \includegraphics[width=\linewidth]{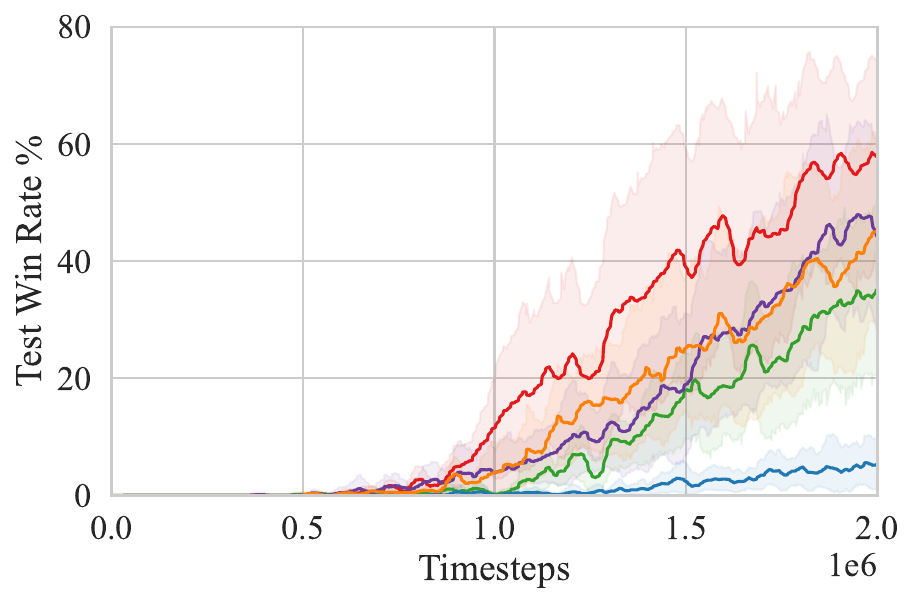}
    \caption{MMM3}
\end{subfigure}
\begin{subfigure}[t]{.32\linewidth}
    \centering
    \includegraphics[width=\linewidth]{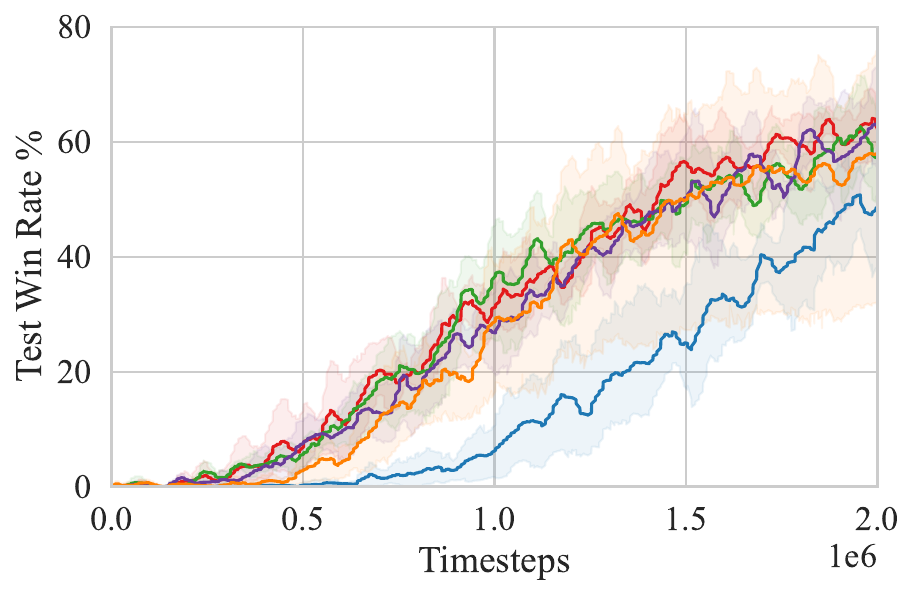}
    \caption{1c3s5z\_vs\_1c3s6z}
    \label{fig:1c}
\end{subfigure}
\begin{subfigure}[t]{.32\linewidth}
    \centering
    \includegraphics[width=\linewidth]{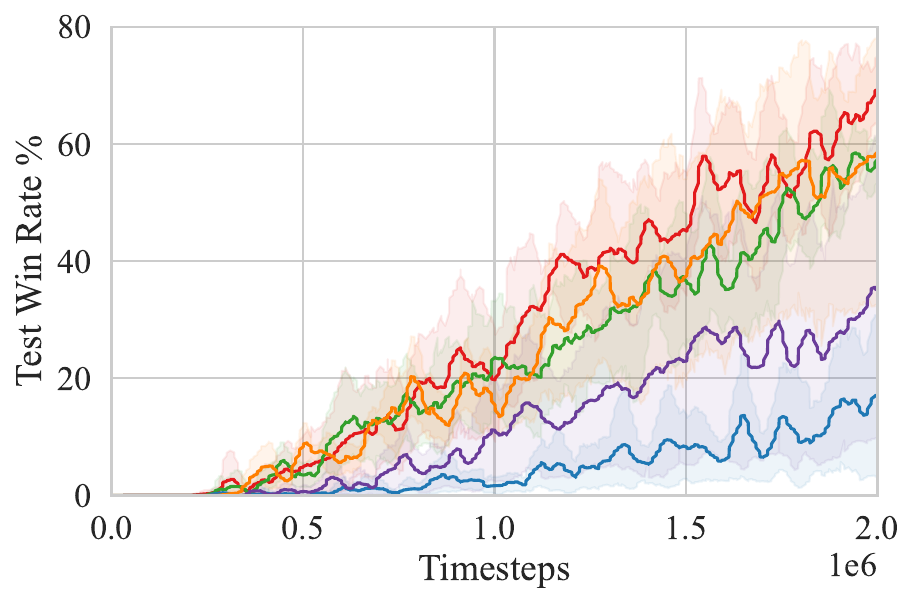}
    \caption{27m\_vs\_30m}
    \label{fig:27m}
\end{subfigure}
\begin{subfigure}[t]{.32\linewidth}
    \centering
    \includegraphics[width=\linewidth]{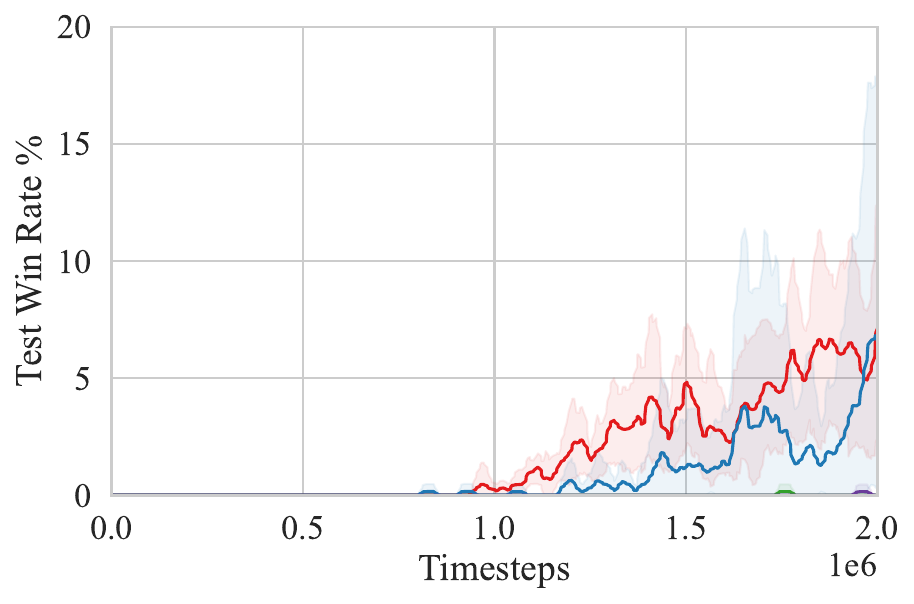}
    \caption{corridor}
\end{subfigure}
\begin{subfigure}[t]{.32\linewidth}
    \centering
    \includegraphics[width=\linewidth]{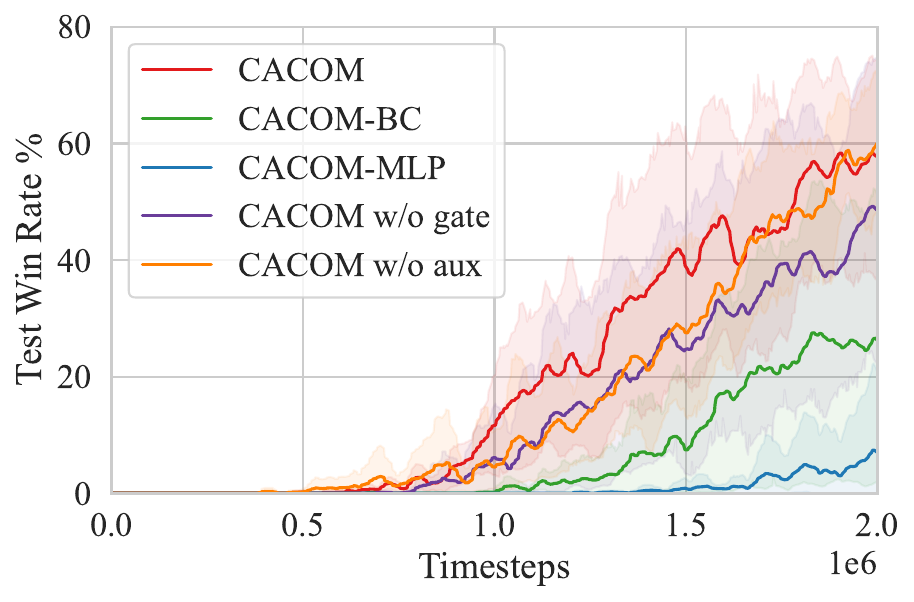}
    \caption{Ablation studies on MMM3}
    \label{fig:abl}
\end{subfigure}
\begin{subfigure}[t]{.32\linewidth}
    \centering
    \includegraphics[width=\linewidth]{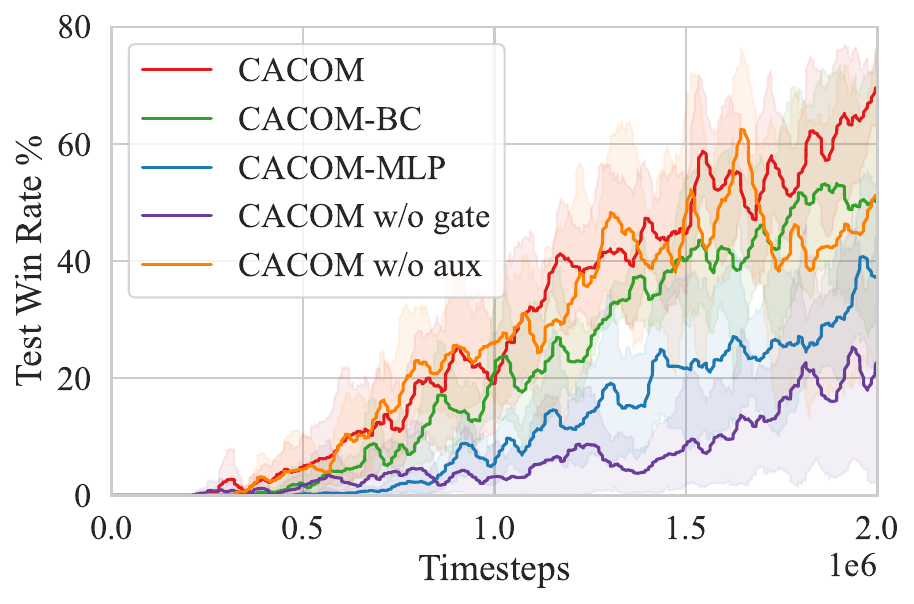}
    \caption{Ablation studies on 27m\_vs\_30m}
    \label{fig:abl_2}
\end{subfigure}

\caption{Performance comparison with baselines on SMAC benchmarks.
}
\label{fig:results_smac}
\Description{Performance comparison with baselines on SMAC benchmarks where CACOM achieves the best performance overall.
}
\end{figure*}

\begin{table}[h]
\centering
\caption{Gate Pruning Results for SMAC}
\begin{tabular}{ccc}
\toprule
scenarios & \makecell{second-stage links \\ pruned ratio} & \makecell{communication budget \\ occupied ratio} \\ \midrule
MMM3  &  $60.52(\pm 13.23) \%$ &  $58.59(\pm 6.19) \%$ \\ 
1c3s5z\_vs\_1c3s6z &  $57.36(\pm 18.10) \%$ &  $56.98(\pm 10.18) \%$ \\ 
27m\_vs\_30m &  $49.96(\pm 4.66) \%$ &  $66.69(\pm 2.07) \%$ \\ 
corridor &  $23.05(\pm 4.60) \%$ &  $84.63(\pm 2.04) \%$ \\ 
\bottomrule
\end{tabular}
\label{table: gate}
\end{table}

\textbf{Results:}
We show the performance of CACOM and baselines on SMAC benchmarks in Figure~\ref{fig:results_smac} and the corresponding gate pruning results in Table~\ref{table: gate}. Overall, CACOM consistently outperfoms the baselines. Compared to the MPE, the observations in SMAC are more sophisticated therefore exacerbate the consequences of limited bandwidth. As a result, most baseline communication protocols perform even worse than the communication-free method QMIX under severe bandwidth constraints (e.g., Figure~\labelcref{fig:1c,fig:27m}). In contrast, CACOM still manages to obtain evident performance gains. This highlights its ability to leverage the limited bandwidth efficiently by transmitting only the most relevant information.

\begin{figure}
  \includegraphics[width=0.7\linewidth]{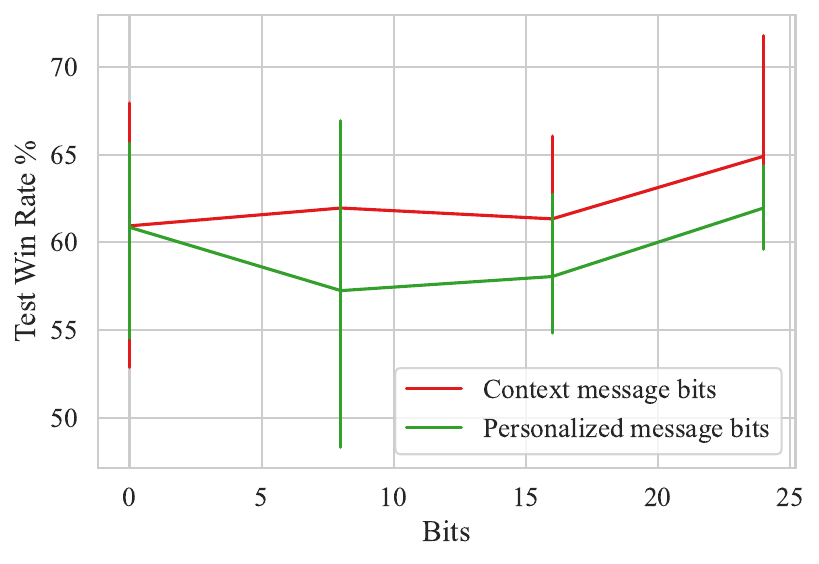}
  \centering
  \caption{Message size analysis. 
  The \textcolor[HTML]{33A02C}{green} curve shows the influence of personalized message bits while the context message is fixed as 8 bits and the \textcolor[HTML]{E31A1C}{red} curve shows the influence of context message bits while the personalized message is fixed as 24 bits.
  }
  \label{fig:msg_len}
  \Description{Message size analysis. 
  The \textcolor[HTML]{33A02C}{green} curve shows the influence of personalized message bits while the context message is fixed as 8 bits and the \textcolor[HTML]{E31A1C}{red} curve shows the influence of context message bits while the personalized message is fixed as 24 bits.
  }

\end{figure}
\textbf{The Effect of Messages Length:}
In Figure~\ref{fig:msg_len}, we demonstrate how the number of bits for the context message and personalized message influence the performance. From the red curve, we see that increasing context message bits nearly never impedes the training because a longer context message enables better personalized message generation. However, increasing the message size for the first stage may result in heavy communication overhead since the first-stage communication operates in a broadcast fashion.

On the other hand, a very small personalized message size can result in worse performance compared to the case without communication, as the green curve suggests. This may be because personalized messages that are supposed to assist the receivers' decision making fail to provide helpful information under such a low communication budget. And this also implies that CACOM may be further improved when we combine it with more delicate design at the receiver side \citep{masis_guan2022efficient}. 

\textbf{Ablation Studies:} To verify the effectiveness of different design choices in CACOM, we conduct ablation studies on MMM3 and 27m\_vs\_30m scenarios, which are illustrated in Figure~\labelcref{fig:abl,fig:abl_2}. We keep the CACOM network architecture but replace the context-aware communication with broadcasting communication and denote it as CACOM-BC. We replace the attention blocks in the network with multi-layer perceptron (MLP) blocks and denote it as CACOM-MLP. We remove the gating mechanism in CACOM and call it CACOM w/o gate. We remove the auxiliary task and call it  CACOM w/o aux. Figure~\labelcref{fig:abl,fig:abl_2} show that CACOM outperforms CACOM-BC, indicating the better expressiveness achieved by context-aware communication. The unsatisfactory performance of CACOM-MLP implies that the success of CACOM shall also attribute to the network design. Without the attention mechanism to help identifying the related information, it is difficult for the helpers to learn to generate the personalized messages. Furthermore, the gating mechanism improves the performance by filtering out the unrelated information and the auxiliary task assists the training by stabilizing the training in some scenarios.

\textbf{Limitations:} While our method has demonstrated its effectiveness in the above experiments, we have observed that the results of gate-pruning exhibit a relatively large variance (especially in MPE). This suggests that the gate training process may be subject to instability.  \looseness=-1

\section{Conclusions}

In this paper, we investigated context-aware communication in  communication-constrained MARL. In an endeavor to optimize the limited communication budget, we proposed CACOM as a context-aware communication protocol, advocating a shift from sender-centric broadcasting communication schemes to receiver-centric personalized communication. Particularly, we utilized various attention-based blocks to discern the information beneficial to receivers and employed LSQ blocks to quantize the messages to be transmitted. Empirical results underscore the effectiveness of CACOM in cooperative MARL under low communication budgets. 
For future research, it is interesting to further enhance the context-aware communication protocols by investigating link scheduling and utilizing historical communication messages to reduce communication overhead. \looseness=-1

\begin{acks}
This work was supported in part by NSFC/RGC Collaborative Research Scheme grant CRS\_HKUST603/22 and in part by the Hong Kong Research Grants Council under the Research Impact Fund (R5009-21).
\end{acks}


\balance
\bibliographystyle{ACM-Reference-Format} 
\bibliography{ref}


\newpage
\appendix
\section{Details of MARL Algorithms} \label{sec: supp_bg}

MADDPG~\citep{maddpg_lowe2017multi} is an actor-critic algorithm that has been commonly adapted for the CTDE paradigm, where a centralized critic $Q\left(\boldsymbol{o}, \boldsymbol{h}, \boldsymbol{a}; \theta \right) $ and individual deterministic policies $\pi_i(o_i, h_i; \theta)$ are learned during the training process ($\theta$ denotes the parameters for deep neural networks). The gradient for the policy network for the $i$-th agent is computed as 
\begin{align}
    \nabla_{\theta_i}J(\theta_i) = &\mathbb{E}_{(\boldsymbol{o}, \boldsymbol{h}, \boldsymbol{a}, r, \boldsymbol{o}', \boldsymbol{h}') \sim \mathcal{D}}  \left[\nabla_{\theta_i}\pi (o_i, h_i; \theta_{i}) \right. \nonumber\\
    &\left.\cdot \nabla_{a_i}Q(\boldsymbol{o}, \boldsymbol{h}, \boldsymbol{a}_{-i}, a_i ; \theta)|_{a_{i}=\pi_i(o_i, h_i; \theta_i)} \right],
\end{align}
while the centralized critic is updated using the temporal difference (TD) learning objective
\begin{align}
    \mathcal{L}_{\text{TD}}(\theta) = \mathbb{E}_{(\boldsymbol{o}, \boldsymbol{h}, \boldsymbol{a}, r, \boldsymbol{o}', \boldsymbol{h}') \sim \mathcal{D}} \left[\left(y - Q\left(\boldsymbol{o}, \boldsymbol{h}, \boldsymbol{a}; \theta \right) \right)^2 \right],
\end{align}
where
$
y = r + \gamma Q\left(\boldsymbol{o}, \boldsymbol{h}', \boldsymbol{a}' ; \theta^-\right)|_{a_{i}' {=}\pi_i(o_i', h_i'; \theta_i^-)},
$ 
and $\theta^-$ denotes the target networks with delayed parameters. 

QMIX~\citep{qmix_rashid2020monotonic} is a multi-agent value-based algorithm that exploits the idea of value decomposition. In this algorithm, the global value function $Q_{\text{tot}}(\boldsymbol{o}, \boldsymbol{h}, \boldsymbol{a}; \theta)$ is decomposed into individual value functions $Q_{i}(o_i, h_i, a_i; \theta_i)$ following the monotonic principle. The training objective is to minimize the TD error, which is computed as
\begin{align}
    \mathcal{L}_{\text{TD}}(\theta) = \mathbb{E}_{(\boldsymbol{o}, \boldsymbol{h}, \boldsymbol{a}, r, \boldsymbol{o}', \boldsymbol{h}') \sim \mathcal{D}} \left[ \left(y - Q_{\text{tot}}(\boldsymbol{o}, \boldsymbol{h}, \boldsymbol{a}; \theta) \right)^2 \right], 
\end{align}
where 
$
    y = r + \gamma \max_{\boldsymbol{a}'} Q_{\text{tot}}(\boldsymbol{o}', \boldsymbol{h}', \boldsymbol{a}'; \theta^-).
$

\section{Implementation Details} \label{sec: supp_impl}
We summarize the execution phase of CACOM from agent $i$'s perspective in Algorithm~\ref{algo_execution}.

\begin{algorithm*}
\caption{Execution Phase of CACOM at Agent $i$}
\begin{algorithmic}[1]
\State \textbf{Inputs:} Local observation $o_i$

\State Local Encoder takes local observation $o_i$ as input and outputs local feature $f_i$ and context message $c_i$
\State Agent $i$ broadcasts it context message $c_i$ to all other agents 
\Comment{The first communication stage, agent $i$ acting as a helpee}

\State Agent $i$ receives context messages $\{c_j\}_{N-1}$ from other agents 
\Comment{The second communication stage, agent $i$ acting as a helper}
\For{$j = 1, 2, ..., N$} 
\If{$j == i $}
\state Continue
\Else 
\State Local gate $\mathcal{G}$ decide whether to help agent $j$ based on $z_{ij}$ following Equation~\ref{eq:local_gate}
    \If {$z_{ij} < 0.5$}
    \state Continue
    \Else
    \State Generate personalized message $m_{ij}$ following Equation~\ref{eq:reply}
    \State Send personalized message $m_{ij}$ to agent $j$
    \EndIf
\EndIf
\EndFor

\State Agent $i$ receives personalized messages $\{m_{ji}\}_{M}$ from other agents 
\Comment{The second communication stage, agent $i$ acting as a helpee}\\
\Comment{$M$ is the number of personalized messages agent $i$ receives}
\State Agent $i$ aggregates the received messages and its local history to obtain local q value $Q_i(o_i, h_i, \cdot; \theta)$ following Equation~\ref{eq: aggregation}

\State \textbf{Outputs:} Local q value $Q_i(o_i, h_i, \cdot; \theta)$

\end{algorithmic}
\label{algo_execution}
\end{algorithm*}


In MPE, both CACOM and baselines are trained based on MADDPG. The centralized critic is realized by three FC layers for all methods. Individual policy networks are realized according to the implementation from the respective original papers while keeping the number of parameters roughly the same across all communication algorithms. Adam optimizer is used with the learning rate tuned for different algorithms. Key hyperparameters shared across all algorithms are shown in Table \ref{table: param_MPE}.
\begin{table}[h]
\centering
\caption{Hyperarameters for MPE}
\begin{tabular}{cc}
\toprule
hyperparameters & value \\ \midrule
discount factor $\gamma$ &  $0.95$  \\ 
gradient clip norm &  $1.0$  \\ 
soft update parameter $\tau$ &  $0.01$  \\
learning interval &  $100$  \\ 
batch size &  $100 \times \textrm{num\_agents}$  \\ 
replay buffer size &  $1,000,000$  \\ 
critic learning rate &  $0.01$  \\ 
\bottomrule
\end{tabular}
\label{table: param_MPE}
\end{table}

We tune the learning rate from $[0.001, 0.0001]$ for all the baselines and list the chosen actor learning rate in Table~\ref{table: param_MPE_lr} . Hyperparameters in CACOM are listed in Table~\ref{table: param_MPE_CACOM}.
\begin{table}[h]
\centering
\caption{Actor learning rates for MPE}
\begin{tabular}{ccc}
\toprule
\multirow{2}{*}{methods} & \multicolumn{2}{c}{actor learning rate} \\  
\cmidrule{2-3}
 & PP & CN  \\ \midrule
MADDPG &  $0.001$ & $0.001$ \\ 
TarMAC + LSQ &  $0.001$ & $0.0001$  \\
I2C + LSQ &  $0.001$ & $0.0001$ \\ 
MAIC + LSQ &  $0.001$ & $0.0001$  \\ 
CACOM & $0.001$ & $0.001$ \\
\bottomrule
\end{tabular}
\label{table: param_MPE_lr}
\end{table}

\begin{table}[h]
\centering
\caption{Hyperarameters in CACOM for MPE}
\begin{tabular}{cc}
\toprule
hyperparameters & value \\ \midrule
\multirow{2}{*}{gate learning interval} 
 & PP: $50,000$ \\ 
 & CN: $100,000$ \\
gate learning start &  $8,000,000$  \\
gate learning rate &  $0.0001$  \\ 
gate threshold $T$ &  $0$  \\ 
\bottomrule
\end{tabular}
\label{table: param_MPE_CACOM}
\end{table}

In SMAC, we adopt all the hyperpameters in PyMARL \citep{smac_samvelyan19smac} except the batch size in the 27m\_vs\_30m scenario. We use 16 for all the baselines and CACOM, instead of 32, due to the limitation of our GPU memory. Compared to the success rate of QMIX reported in the original paper \citep{smac_samvelyan19smac}, we do not see any performance degradation owing to this change. Hyperparameters in CACOM are listed in Table~\ref{table: param_SMAC_CACOM}. \looseness=-1

\begin{table}[h]
\centering
\caption{Hyperarameters in CACOM for SMAC}
\begin{tabular}{cc}
\toprule
hyperparameters & value \\ \midrule
\multirow{2}{*}{gate learning interval} 
 & corridor: $1,000,000$ \\ 
 & others: $10,000$ \\
gate learning start &  $200,000$  \\
gate learning rate &  $0.0001$  \\ 
gate threshold $T$ &  $0$ \\ 
\bottomrule
\end{tabular}
\label{table: param_SMAC_CACOM}
\end{table}

We start gate training for corridor way later than other scenarios because this scenario is more challenging than others. Cutting off links in an early stage will result in unsatisfactory performance.

Random seeds 0, 1, 2, 3, 4 are used for different runs. All the experiments are conducted on NVIDIA GeForce RTX 3080 GPUs.

\section{Environmental Settings}\label{sec: supp_env}

\begin{figure*}[t]
\centering
\begin{subfigure}[c]{.37\linewidth}
    \centering
    \includegraphics[width=\linewidth]{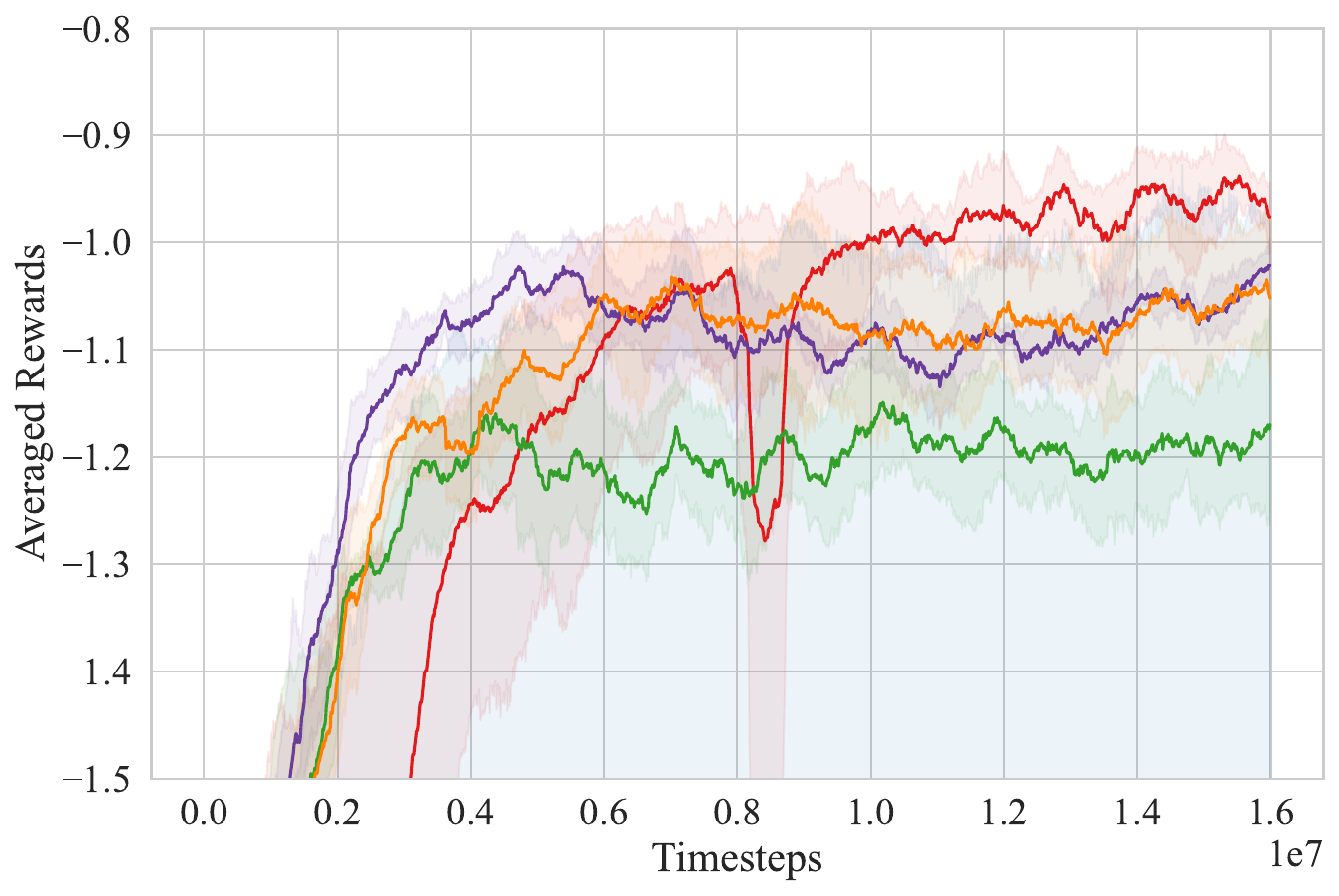}
    \caption{16 bits per link}
    \label{fig:mpe_pp_16}
\end{subfigure}
\begin{subfigure}[c]{.37\linewidth}
    \centering
    \includegraphics[width=\linewidth]{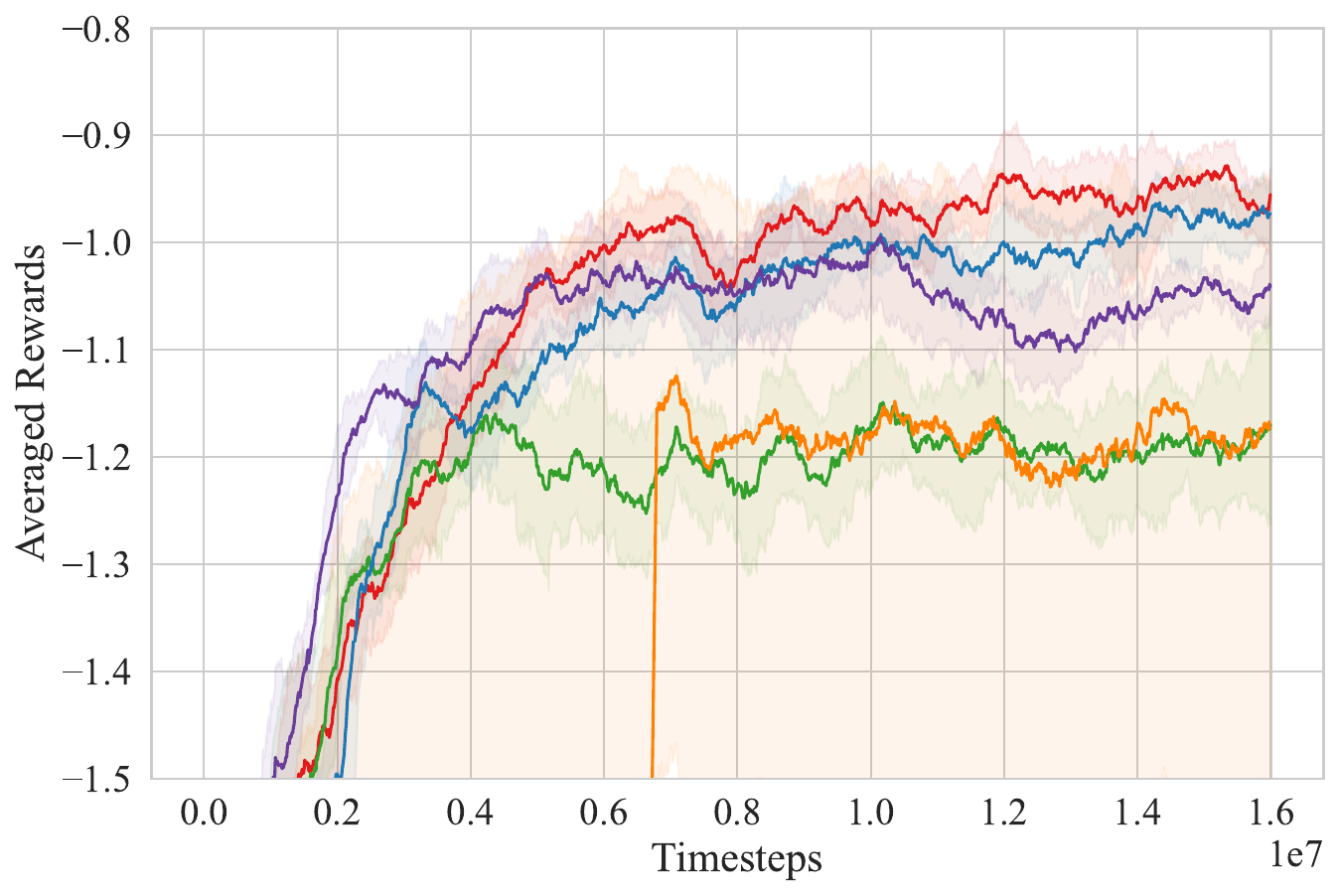}
    \caption{32 bits per link}
    \label{fig:mpe_pp_32}
\end{subfigure}
\begin{subfigure}[c]{.15\linewidth}
    \centering
\includegraphics[width=\linewidth]{figs/MPE_legend_cropped_0504.pdf}

\end{subfigure}

\caption{Performance comparison with different communication budgets on PP.
}
\label{fig:results_mpe_different_comm}
\end{figure*}

For MPE implementations, we follow I2C \citep{i2c} except for altering the number of agents, preys and landmarks to make the tasks more challenging.
In PP, the team reward is calculated by taking the negative sum of the distances between all preys and their closest corresponding predators. Preys move faster than predators and are preprogrammed to move away from the nearest predator, but the preys are outnumbered by the predators. Therefore, it is essential for agents to utilize such an advantage and act collaboratively. Moreover, since each agent is only granted partial observation and the communication budget is limited, it is crucial for the communication messages to be concise. 

In CN, the team reward is calculated by taking the negative sum of the distances between all landmarks and their closest corresponding agents. Agents can only observe the landmarks close to them, thus they need to rely on communication to obtain knowledge of vacant landmarks observed by other agents. In this case, it is crucial to convey precise messages that can aid the helpee in making informed decisions.

We list the environment parameters in Table~\ref{table: param_MPE_env}. 

\begin{table}[h]
\centering
\caption{Environment parameters in MPE}
\begin{tabular}{ccc}
\toprule
\multirow{2}{*}{parameters} & \multicolumn{2}{c}{value} \\  
\cmidrule{2-3}
 & PP & CN  \\ \midrule
number of agents &  $10$ & $8$ \\ 
number of preys/landmarks &  $4$ & $8$  \\
number of observed agents &  $2$ & $4$ \\ 
number of observed preys/landmarks &  $2$ & $4$  \\ 
world dimension & \multicolumn{2}{c}{$2$} \\
agent size & \multicolumn{2}{c}{$0.05$} \\
agent acceleration & \multicolumn{2}{c}{$5.0$} \\
prey acceleration & $7.0$ & - \\
random initial locations & \multicolumn{2}{c}{True} \\
reward captures & \multicolumn{2}{c}{False} \\
collision penalty & \multicolumn{2}{c}{$-1$} \\
steps per episode & \multicolumn{2}{c}{$40$} \\
\bottomrule
\end{tabular}
\label{table: param_MPE_env}
\end{table}

\section{Additional Experiments}\label{sec: supp_exp}
We conduct additional experiments on the PP environment with different communication budgets and show the results in Figure~\ref{fig:results_mpe_different_comm}. We observe that although in all the cases (where the communication budgets per link is set to $ [16, 24, 32] $ bits), CACOM outperforms the baselines, the performance gain is more significant when the communication budgets is low. This further demonstrates that our proposed CACOM is effective under communication-constraint scenarios.

\end{document}